\documentclass[letterpaper, 10 pt, conference]{ieeeconf}  

\IEEEoverridecommandlockouts                              

\bstctlcite{IEEEexample:BSTcontrol}

\usepackage{graphicx} 
\usepackage{epsfig} 
\usepackage{mathptmx} 
\usepackage{times} 
\usepackage{amsmath} 
\usepackage{amssymb}  
\usepackage{psfrag}
\usepackage{subcaption}
\usepackage{array} 
\usepackage{soul} 

\usepackage[table,xcdraw]{xcolor}
\usepackage{multirow}

\definecolor{grey_plot}{rgb}{0.6588235294117647, 0.6431372549019608, 0.5843137254901961}

\newcommand{\degree}{\ensuremath{^\circ}}
\newcolumntype{g}{>{\columncolor{grey_plot!30}}c}
\newcolumntype{L}{>{\columncolor{grey_plot!30}}l}

\title{\LARGE \bf
OffsetNet: Deep Learning for Localization in the Lung using Rendered Images  
}

\author{Jake Sganga$^{1*}$,  David Eng$^{2*}$, Chauncey Graetzel$^{3}$ and David Camarillo$^{1}$
\thanks{*These authors contributed equally.}
\thanks{This work was supported by the NIH Biotechnology Training Grant, NVIDIA GPU Grant, and Auris Health, Inc.}
\thanks{$^{1}$Department of Bioengineering, Stanford University, Stanford, CA 94305}%
\thanks{$^{2}$Department of Computer Science, Stanford University,  Stanford, CA 94305}%
\thanks{$^{3}$Auris Health Inc.,  Redwood City, CA 94065}%
\thanks{{\tt\small \{sganga,dkeng,dcamarillo\}@stanford.edu}}
\thanks{{\tt\small chauncey.graetzel@aurishealth.com}}
}

\begin{document}

\captionsetup[table]{skip=0pt}        
\addtolength{\parskip}{-0.5mm}

\maketitle
\thispagestyle{empty}
\pagestyle{empty}

\begin{abstract}
Navigating surgical tools in the dynamic and tortuous anatomy of the lung's airways requires accurate, real-time localization of the tools with respect to the preoperative scan of the anatomy. Such localization can inform human operators or enable closed-loop control by autonomous agents, which would require accuracy not yet reported in the literature. In this paper, we introduce a deep learning architecture, called OffsetNet, to accurately localize a bronchoscope in the lung in real-time. After training on only 30 minutes of recorded camera images in conserved regions of a lung phantom, OffsetNet tracks the bronchoscope's motion on a held-out recording through these same regions at an update rate of 47 Hz and an average position error of 1.4 mm. Because this model performs poorly in less conserved regions, we augment the training dataset with simulated images from these regions. To bridge the gap between camera and simulated domains, we implement domain randomization and a generative adversarial network (GAN). After training on simulated images, OffsetNet tracks the bronchoscope's motion in less conserved regions at an average position error of 2.4 mm, which meets conservative thresholds required for successful tracking. 
\end{abstract}

\section{Introduction}
\begin{figure}[thpb]
  \centering
  \includegraphics[width=0.5\textwidth]{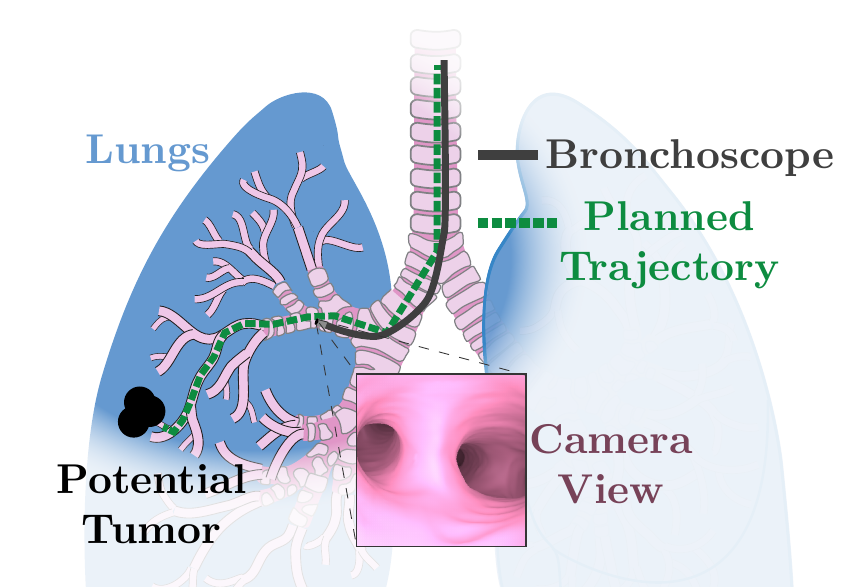}
  \caption{The objective of a bronchoscopy is to navigate the bronchoscope to the site of a potential tumor given the trajectory from the preoperative CT scan. To do this, the operator must use the 2D images from the bronchoscope to determine where it is along the 3D path to the target.}
  \label{lung_task}
  \vspace*{-1mm}
\end{figure}

Early diagnosis of lung cancer, the leading cause of cancer death, significantly improves patient outcomes \cite{DelaCruz2011}. While nodules in the lung's periphery can be diagnosed through thoroscopic surgery or needle biopsy, bronchoscopies are the preferred approach given the lower complication rates (2.2\% vs 20.5\%) \cite{Ost2016,DiBardino2015}. In bronchoscopy procedures, physicians manually drive long, flexible bronchoscopes through the patient's airways to biopsy potentially cancerous nodules, shown in Fig. \ref{lung_task}. Physicians rely on sensor feedback from an on-board camera and, in navigated bronchoscopy procedures, an electromagnetic position sensor at the distal tip of the device. The position sensor is registered to a preoperative computed tomography (CT) of the patient's chest to provide a road map to the target site \cite{Rosell2012}. However, there is significant variability in the diagnostic yield among institutions ranging from 67-74\% \cite{Ost2016,Khan2011}. Accurate localization and robotic control of bronchoscopes can alleviate this variability and improve patient outcomes. 

Since the physician identifies target nodules in the CT reference frame  before the operation begins, the physician must map the sensor feedback from the device (2D image) to the CT frame (3D map). This process is called localization. A localization module must inform the physician how the bronchoscope's current location relates to its desired location. An autonomous agent could use this same information to ``drive'' the bronchoscope without human intervention.

We set out to design an image-based system that can localize in real-time and accurately enough to be used for closed-loop control in a robotic system. We consider image-based approaches instead of hybrid approaches that integrate information from electromagnetic position sensors because these sensors experience unknown biological motion of the patient, have the potential for noise and distortion from metal in the operating room, and increase the total cost of the system \cite{Reynisson2014}. 

Using traditional computer vision techniques to solve this task accurately and in real-time has proven challenging. Several groups have compared the images from the bronchoscope to simulated images rendered from the estimated location of the bronchoscope in CT frame; however, these methods register images inefficiently at around 1-2 Hz with high average registration errors of 3-5 mm \cite{Rai2008,Mori2001,Reynisson2014}. Tracking features using methods like SIFT and ORBSLAM have been used, but the airways have insufficient features and tracked features often drop out \cite{Byrnes2014,Visentini-Scarzanella2017}. Anatomical landmarks have been tracked, like bifurcations \cite{Shen2017}, lumen centers \cite{sanchez2017towards}, centerline paths \cite{Hofstad2014}, or similar image regions \cite{Luo2014}, but these approaches make assumptions about the airway geometries and struggle with image artifacts. 

Merritt \textit{et al.} describes a real-time localization approach with average errors as low as 1.4 mm in simulation \cite{Merritt2013}. Their method precomputes image gradients of simulated reference images along the anticipated procedure path and uses an iterative Gauss-Newton gradient-descent to determine the transformation between each camera image and the closest of these reference images. While this technique reports continuous tracking, it relies on high-quality rendering and a dense collection of reference images.

Because of the difficulties traditional computer vision techniques face in this task, we decided to explore a deep learning approach. Using convolutional neural networks (CNN) to estimate the position and orientation of objects has been shown in many contexts, including for human posture and objects in a hand \cite{Zhou_2016_CVPR,OpenAI2018}.  Visentini-Scarzanella \textit{et al.} used a CNN to estimate the depth map of 2D images in a lung phantom, which could then be registered to the 3D map, but tracking is not reported \cite{Visentini-Scarzanella2017}.

In this work, we contribute an image-based deep-learning approach, called OffsetNet, that localizes a bronchoscope in the CT frame accurately and in real-time. We evaluate OffsetNet on a recorded trajectory in a lung phantom, demonstrating continuous, real-time tracking. We also show that training on simulated images can improve the performance in regions of the lung without recorded training data. 

\vspace*{2mm}

\begin{table}[h]
\caption{Notation}
\label{notation_table}
\centering
\begin{tabular}{>{\centering\arraybackslash}p{0.65cm} p{7.05cm}}
\hline
\multicolumn{2}{c}{\textbf{Input Images into OffsetNet (Fig. \ref{ct_overview})}}
\\
 \rowcolor[HTML]{EFEFEF}
 $\mathbf{I}_{x}^{sty}$ & Generically, an image with style $sty$ at 6-DOF location,  $x$
 \\
 $\mathbf{I}_{x_t}$ & Image from the bronchoscope's current location, $x_t$
 \\
 $\mathbf{I}_{\hat{x}_{t-1}}$ & Image rendered at the previously estimated location, $\hat{x}_{t-1}$
 \\
\hline
\multicolumn{2}{c}{\textbf{Image Styles (Fig. \ref{lung_pics})}}
 \\
 \rowcolor[HTML]{EFEFEF}
 $\mathbf{I}^{\text{cam}}$ & Image taken by a bronchoscope in the lung phantom 
 \\
 $\mathbf{I}^{\text{sim}}$ & Image rendered by OpenGL using the lung CT
 \\ 
 \rowcolor[HTML]{EFEFEF}
 $\mathbf{I}^{\text{rnd}}$ & Image rendered by OpenGL using the lung CT with varied rendering parameters and varied noise, smoothing and occlusions added \cite{Tobin2017}
 \\ 
 $\mathbf{I}^{\text{gan}}$ & Image rendered by OpenGL using the lung CT, then passed through the generator of a trained GAN (generative adversarial network) \cite{Zhu2017e}
 \\              
\hline
\multicolumn{2}{c}{\textbf{Error between True and Estimated Locations}}
\\
 \rowcolor[HTML]{EFEFEF}
$e_p$ & Position error (mm), defined as $e_p$ in \cite{Merritt2013}
\\   
$e_d$ & Direction angle error between pointing vectors, $p_z$, of the two views (\degree), defined as $e_d$ in \cite{Merritt2013}
\\
\rowcolor[HTML]{EFEFEF}
$e_r$ & Roll angle error between the $p_x$ axis after the $e_d$ was corrected for between views (\degree), defined as $e_r$ in \cite{Merritt2013}
\\              
$\|\cdot\|_E$ & Location error, $e_p + 0.175e_d +0.175e_r$, (mm,\degree)
\\              
\hline
\multicolumn{2}{c}{\textbf{Lung Regions (Fig. \ref{training_dist})}}
\\
 \rowcolor[HTML]{EFEFEF}
seen,  $S$ & Airways in the lung where $\mathbf{I}^{\text{cam}}$ images were included in the training set, representing conserved regions among patients
\\   
unseen, $U$ & Airways in the lung where $\mathbf{I}^{\text{cam}}$ images were omitted from the training set, representing less conserved regions
\\
\hline
\end{tabular}
\end{table}
\begin{figure}[thpb]
      \centering
      \includegraphics[width=0.45\textwidth]{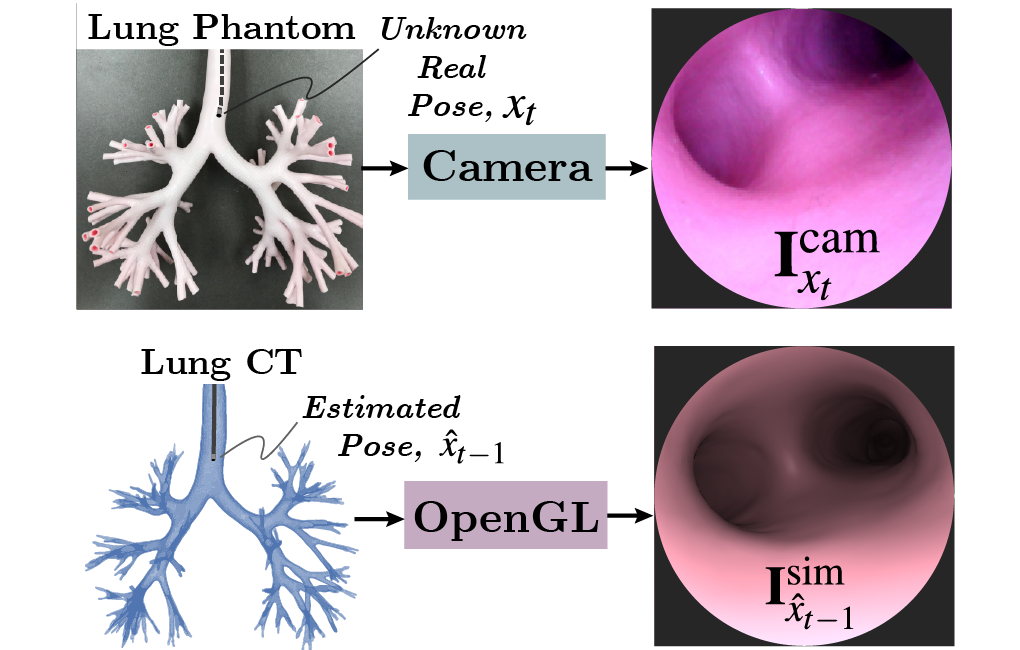}
      \caption{The input to OffsetNet is a $(\mathbf{I}_{x_t}^{\text{cam}}, \mathbf{I}_{\hat{x}_{t-1}}^{\text{sim}})$ image pair, shown here. A bronchoscope inside the lung generates $\mathbf{I}_{x_t}^{\text{cam}}$, and a rendering of the estimated pose in the lung's CT model creates $\mathbf{I}_{\hat{x}_{t-1}}^{\text{sim}}$. The task is to identify the offset between them, and through repeated iterations, reduce the offset as much as possible.}
      \label{ct_overview}
\end{figure}

\begin{figure}[thpb]
      \centering
      \includegraphics[width=0.45\textwidth]{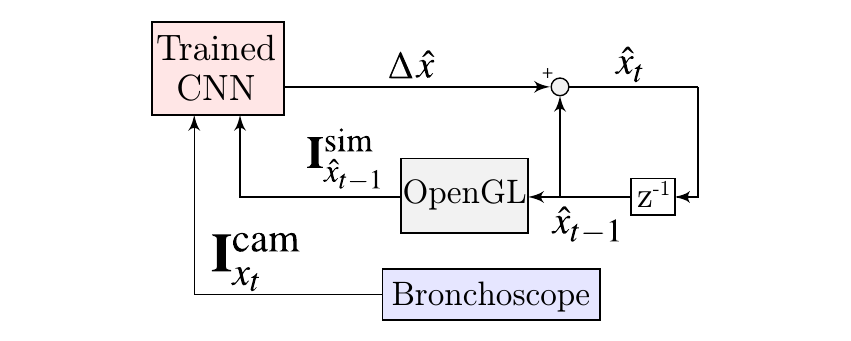}
      \caption{This control loop describes the tracking task, where a trained CNN (OffsetNet) receives a $(\mathbf{I}_{x_t}^{\text{cam}}, \mathbf{I}_{\hat{x}_{t-1}}^{\text{sim}})$ image pair at time $t$ and outputs the estimated 6-DOF offset, $\Delta \hat{x}$.  The estimated location $\hat{x}_t$ is updated, and the updated location is used to render the next $\mathbf{I}_{\hat{x}_{t-1}}^{\text{sim}}$.}
      \label{control_loop_tracking}
\end{figure}

\section{Method}
In this approach, we compare $\mathbf{I}_{x_t}$, the image from the bronchoscope's current location $x_t$, to $\mathbf{I}_{\hat{x}_{t-1}}$, the image rendered at the bronchoscope's estimated location from the previous timestep $\hat{x}_{t-1}$, shown in Fig. \ref{ct_overview}, and estimate the location offset between the images, $\Delta \hat{x}$. This offset updates the estimated bronchoscope location for the current timestep according to $\hat{x}_t \leftarrow \hat{x}_{t-1} + \Delta \hat{x}$, which is used to render the image at the bronchoscope's estimated location in the next time step. By iteratively updating the estimated location, the algorithm can track the motion of the bronchoscope, shown in Fig. \ref{control_loop_tracking}. 

\begin{figure}[thpb]
      \centering
      \includegraphics[width=0.5\textwidth]{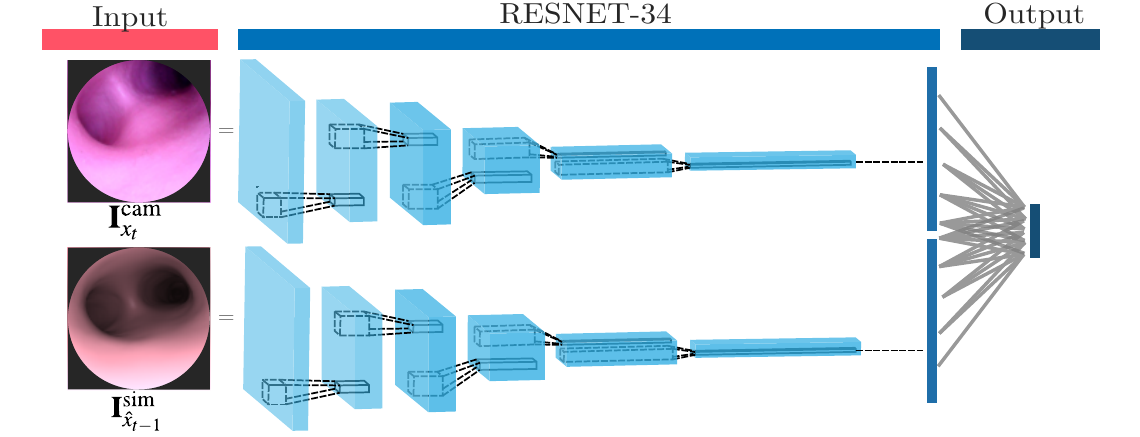}
      \caption{OffsetNet architecture overview show two Resnet-34's processing each image input, followed by a fully connected later, and outputting a 6D vector representing a translation (mm) and rotation (\degree) \cite{He2015}.}
      \label{cnn_architecture}
\end{figure}

Our system consists of two independent deep residual convolutional networks (CNN) with identical architectures, shown in Fig. \ref{cnn_architecture}. For each $( \mathbf{I}_{x_t} , \mathbf{I}_{\hat{x}_{t-1}} )$ image pair, we feed $\mathbf{I}_{x_t}$ into the first network and $\mathbf{I}_{\hat{x}_{t-1}}$ into the second network, concatenate the embeddings, and pass the resulting vector through a fully-connected layer, which produces an estimate of the pose of the first image in the frame of the second image. This length 6 vector, $\Delta \hat{x}$, consists of 3 components describing the position offset $\Delta \hat{x}_p = [p_x, p_y, p_z]$ (mm) and 3 components describing the rotation offset $\Delta \hat{x}_r = [\alpha, \beta, \gamma]$ (\degree), defined by the three Euler angles about axes $xyz$ \cite{craig2005introduction}. We define positive $p_z$ into the page, and positive $p_x$ to the right.

 The residual parts of our network implement the 34-layer architecture described in He \textit{et al.} \cite{He2015}. The CNN was implemented in Tensorflow, version 1.9 \cite{45381}.

\begin{figure}[thpb]
      \vspace*{2mm}
      \centering
      \includegraphics[width=0.5\textwidth]{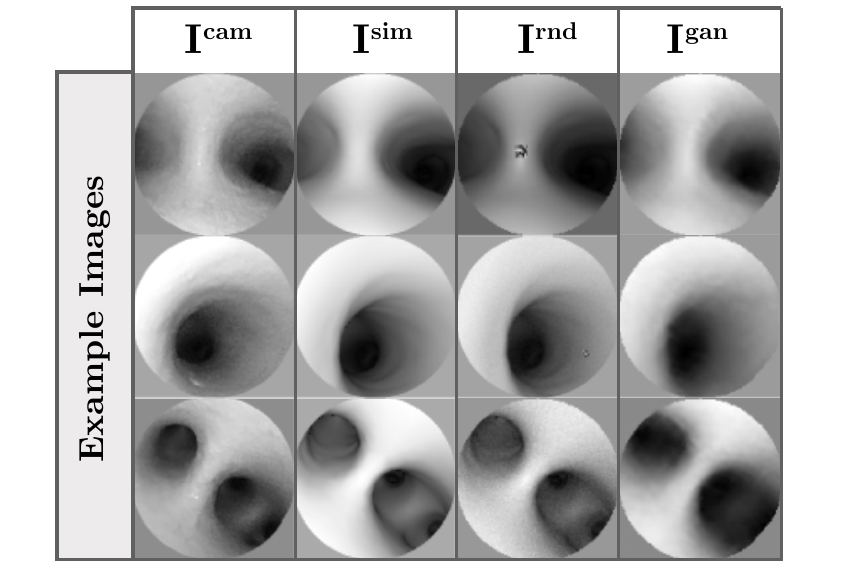}
      \caption{Example images of the lung airways based on image style after grayscaling, adding a circle mask, and per-image normalizing to zero mean and unit standard deviation.}
      \label{lung_pics}
      \vspace*{-2mm}
\end{figure}

The network is trained using Adam optimization to minimize a weighted L2 loss function $\ell(\Delta \hat{x}, \Delta x)$ between the estimated offset, $\Delta \hat{x}$, and the ground-truth offset, $\Delta x$. To relate position and rotation errors, we chose a 1 mm:5.7\degree \ ratio, which roughly relates to the fact that a 5.7\degree \ $e_d$ angle error results in an error of 1 mm for a location 10 mm in front of the camera:
\[
  \ell(\Delta \hat{x}, \Delta x) = ||\Delta \hat{x}_p - \Delta x_p||_2^2 + 0.175 \cdot ||\Delta \hat{x}_r - \Delta x_r||_2^2
\]

When deployed, our control loop compares $\mathbf{I}_{x_t}^{\text{cam}}$, captured by the bronchoscope's camera from its current location, and $\mathbf{I}_{\hat{x}_{t-1}}^{\text{sim}}$, rendered at the bronchoscope's previous estimated location. To reflect this, we evaluate our models on test sets consisting of $( \mathbf{I}_{x_t}^{\text{cam}}, \mathbf{I}_{\hat{x}_{t-1}}^{\text{sim}} )$ image pairs. We train our models on training sets consisting of $( \mathbf{I}_{x_t}^{\{\text{cam,sim,rnd,gan}\}}, \mathbf{I}_{\hat{x}_{t-1}}^{\text{sim}} )$ image pairs, where the style of the image from the bronchoscope's current location varies according to the experiment. Note that all image pairs consist of $(\cdot, \mathbf{I}_{\hat{x}_{t-1}}^{\text{sim}})$.

\begin{figure}[thpb]
      \centering
      \includegraphics[width=0.5\textwidth]{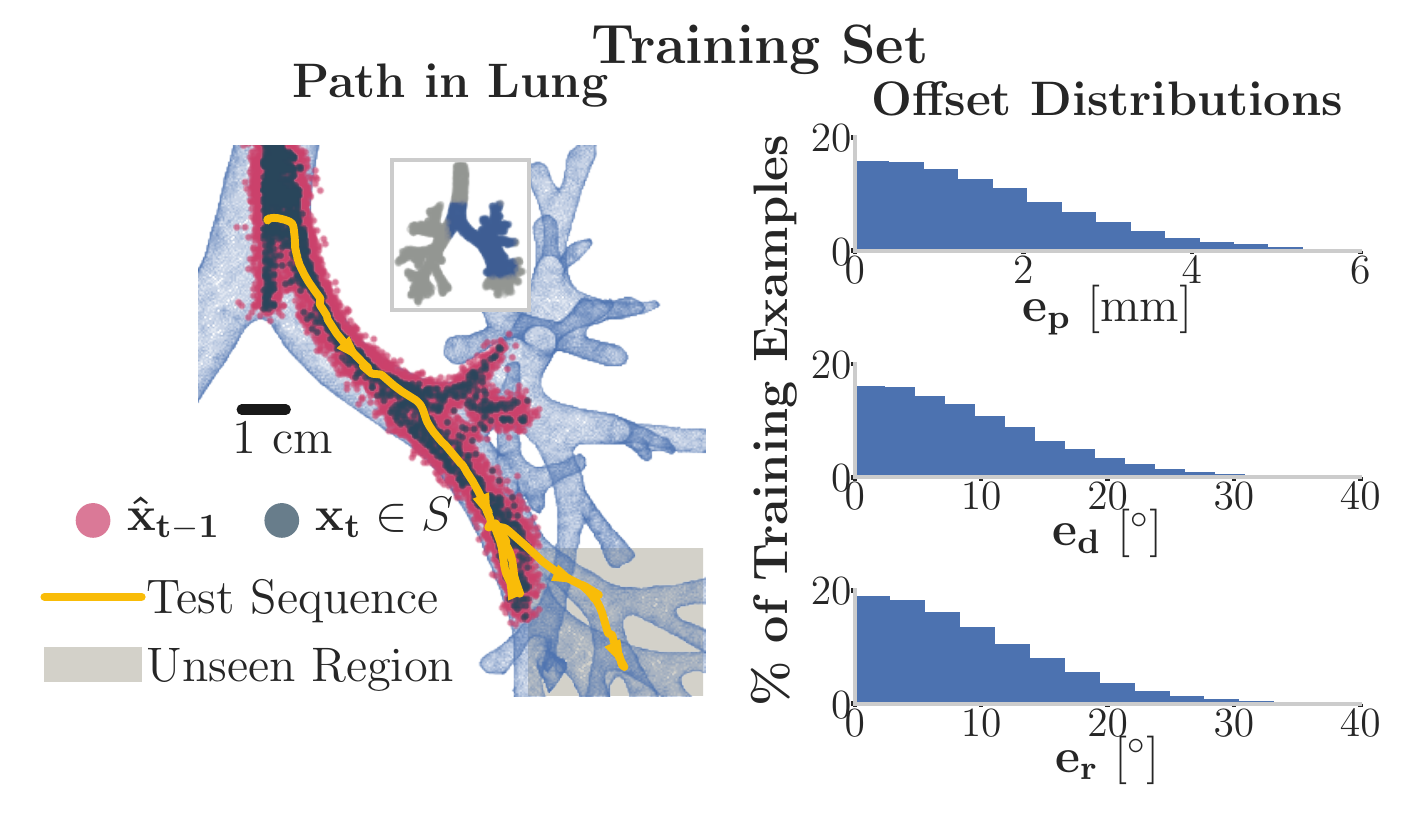}
      \caption{Left, a sample of locations $x_t \in S$ and $\hat{x}_{t-1}$ used in the training set. Notice the unseen region lacks any $x_t$. Right, the distributions of offsets between the $(x_t, \hat{x}_{t-1})$ pairs is shown for $e_p$, $e_d$ and $e_r$.}
      \label{training_dist}
      \vspace*{-1.75mm}
\end{figure}

All $\mathbf{I}^{\text{cam}}$ images were created by manually driving a robotic bronchoscope (Monarch Platform, Auris Health Inc.) for 30 minutes in a lung phantom (Koken Co.), covering 3-5 generations of both lungs. For the experiments presented, OffsetNet was trained on data in the left lung to reduce the data file sizes, shown in Fig \ref{training_dist}. While driving, a 6-DOF electromagnetic sensor (Northern Digital Inc.) tracked the bronchoscope and provided an initial, coarse estimate of the location in CT frame $\tilde{x}_t$ from which each $\mathbf{I}_{x_t}^{\text{cam}}$ was captured. However, a single rigid registration of the sensor's output resulted in errors of up to several millimeters, so local registrations were iteratively refined to improve the label quality. Local registrations were performed by manually translating and rotating nearby points (within a 2 cm cube) until the pixel-wise error $\|\mathbf{I}_x^{\text{sim}} - \mathbf{I}_x^{\text{cam}}\|_2$ reached a local minimum. Powell's method was used in conjunction with manual optimization to refine $\tilde{x}_t$ to a ground-truth location in CT frame $x_t$ \cite{powell1963}.

All $\mathbf{I}^{\text{sim}}$ images are rendered using PyOpenGL \cite{fletcher2005pyopengl} and a 3D lung STL from a segmented CT scan of the lung phantom (Monarch Platform, Auris Health Inc.) \cite{mansoor2015segmentation}. The rendering parameters are based on Higgins \textit{et al.} with a field of view of 60\degree \ \cite{Higgins2008}. Images are rendered at 60 Hz on a PC with no accelerations. The lighting and color were optimized to reduce the pixel-wise difference between images with different styles at the same location, $\mathbf{I}_{x}^{\text{sim}}$ and $\mathbf{I}_{x}^{\text{cam}}$, after grayscaling and per-image normalization. The quality of the segmentation critically affects the quality of $\mathbf{I}^{\text{sim}}$.

Training sets used in Section \ref{unseen_section} also consist of $\mathbf{I}^{\text{rnd}}$ and $\mathbf{I}^{\text{gan}}$ images.

Tobin \textit{et al.} first introduced $\mathbf{I}^{\text{rnd}}$ images, which were rendered with randomized parameters to train their CNN to be indifferent to these changes \cite{Tobin2017}. We randomized each parameter with a normal distribution centered about the default rendering parameters. For these experiments, brightness, attenuation factor, specular intensity, and ambient intensity were all varied by 1, 0.001, 0.1, and 0.1, respectively. In addition, we used randomized Gaussian smoothing, and on half of the images, we added independent per-pixel noise and white noise occlusions of various sizes.

To make our rendered images more closely resemble $\mathbf{I}^{\text{cam}}$, we trained a GAN, following the general design principles described in Zhu \textit{et al.} \cite{Zhu2017e}. We adopted the architecture for our generative network from Johnson \textit{et al.} \cite{Johnson2016}. For the discriminator networks, we used $14 \times 14$ PatchGANs. We encouraged a pixel-wise similarity between the transferred $\mathbf{I}^{\text{cam}}$ and the actual $\mathbf{I}^{\text{cam}}$ images, as well as the transferred $\mathbf{I}^{\text{sim}}$ and the actual $\mathbf{I}^{\text{sim}}$ images in our loss function. We trained on 1000 total $\mathbf{I}^{\text{sim}}$ images and 1000 total $\mathbf{I}^{\text{cam}}$ images. Similar to Zhu \textit{et al.}, we replaced negative log-likelihood objective by a least-squares loss function.

\begin{figure*}[h!] 
      \centering
      \includegraphics[width=1.\textwidth]{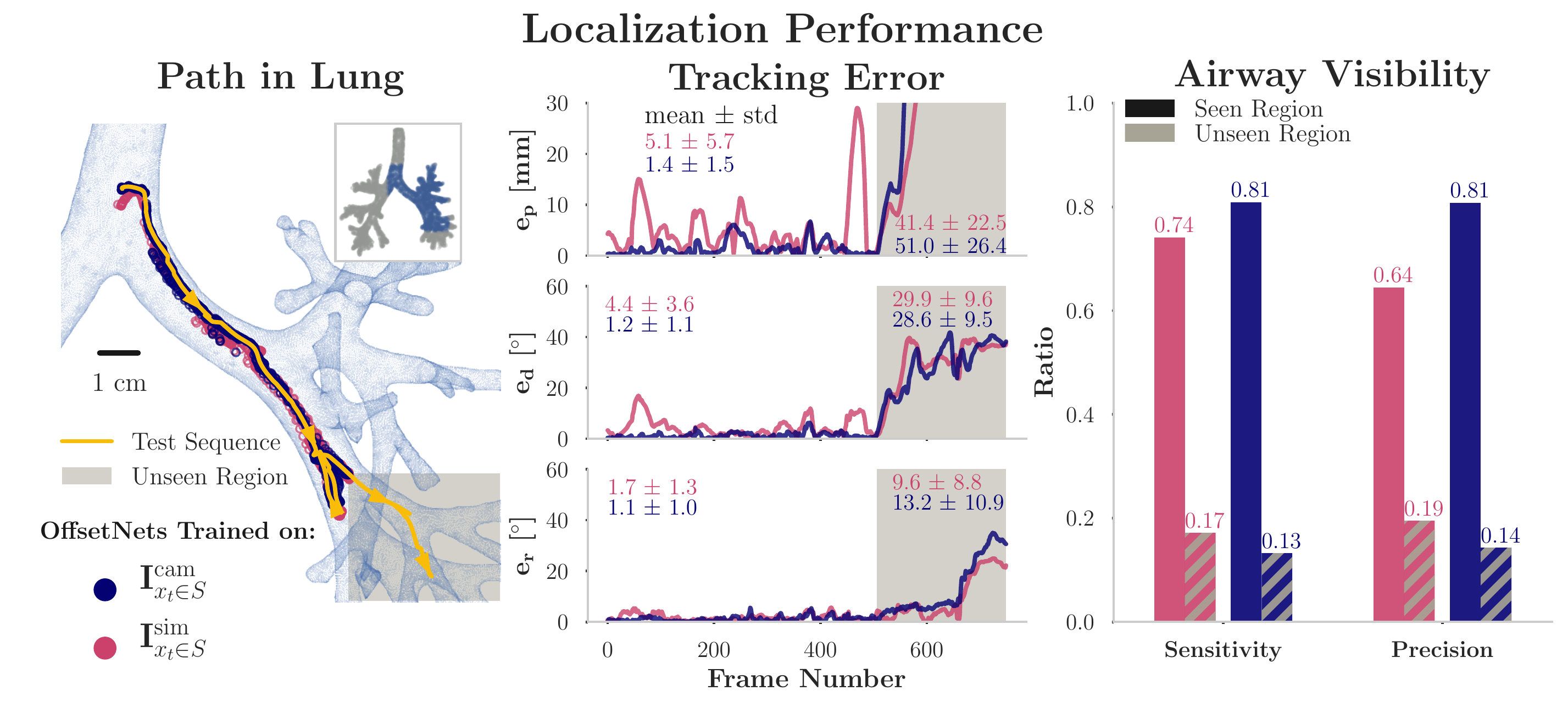}
      \caption{Two OffsetNets are shown in a tracking task on the test sequence, $\mathbf{I}_{x}^{\text{cam}}$. Note that all models were trained on $(\cdot, \mathbf{I}^{\text{sim}}_{\hat{x}_{t-1}})$ image pairs. Left, the path and estimated positions are shown on the lung CT. Middle, the tracking error in $e_p$, $e_d$, and $e_r$ are shown over the frame sequence, and the frames in the unseen region are highlighted in gray. Right, the airway visibility analysis shows sensitivity and precision ratios for the seen and unseen (hatched) regions.}
      \label{koken_tracking}
\end{figure*}

To generate datasets consisting of $( \mathbf{I}_{x_t} , \mathbf{I}_{\hat{x}_{t-1}} )$ image pairs, for each registered $\mathbf{I}_{x_t}$, we generate a $\mathbf{I}_{\hat{x}_{t-1}}$ at a location $\hat{x}_{t-1}$ offset from the registered location $x_t$ according to normal distributions of $e_p$, $e_d$, and $e_r$. The distributions used for the training had 0 mean and standard deviations of 2 mm, 11\degree, 11\degree, respectively, shown in Fig. \ref{training_dist}. Each $\mathbf{I}_{x_t}^\text{cam}$ was augmented by rotating the images with a normal distribution of 0 mean and 14\degree \ standard deviation.

Training sets in Section \ref{seen_section} comprised of 200 $( \mathbf{I}_{x_t}^{\text{cam}}, \mathbf{I}_{\hat{x}_{t-1}}^{\text{sim}} )$ image pairs, where each location $x_t$ was offset (only rolled) from an original recorded location $x_t^\text{rec}$, and each location $\hat{x}_{t-1}$ was offset from the resulting location $x_t$. Training sets in Section \ref{unseen_section} comprised of 750 $( \mathbf{I}_{x_t}^{\{\text{sim,rnd,gan}\}}, \mathbf{I}_{\hat{x}_{t-1}}^{\text{sim}} )$ image pairs, at locations $x_t$ offset from recorded locations $x_t^\text{rec}$ in the test sequence. Each location $\hat{x}_{t-1}$ was offset from the resulting location $x_t$.

\begin{figure}[thpb]
      \vspace*{-3mm}
      \centering
      \includegraphics[width=0.5\textwidth]{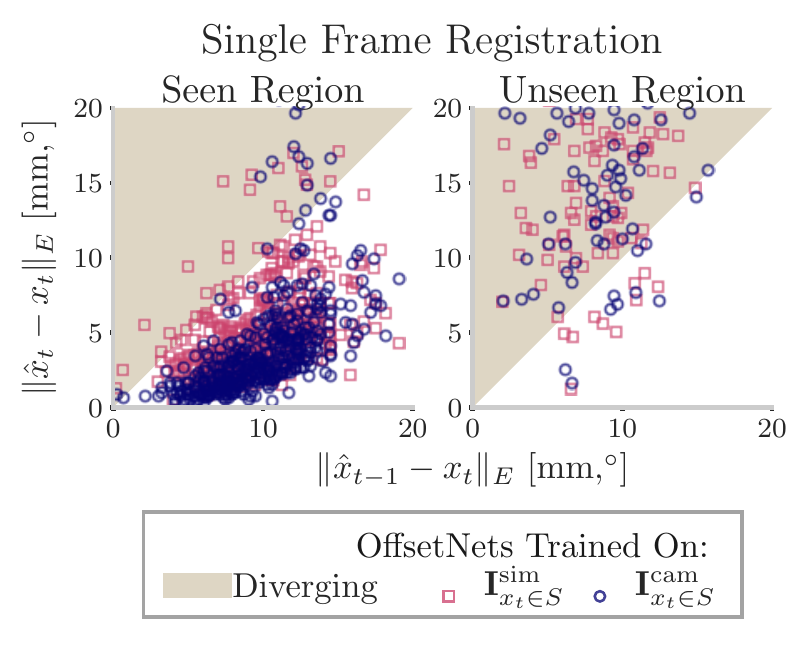}
      \caption{OffsetNets tested on a dataset of $(\mathbf{I}_{x_t}^{\text{cam}}, \mathbf{I}_{\hat{x}_{t-1}}^{\text{sim}})$ image pairs along the full test sequence. 25 $\hat{x}_{t-1}$ locations were randomly offset from each $x_t$. Note that all models were trained on $(\cdot, \mathbf{I}^{\text{sim}}_{\hat{x}_{t-1}})$ image pairs. Each axis is a linear combination of translation and rotation error, defined in Table \ref{notation_table}. Diverging estimates are defined by the $\|\hat{x}_t - x_t\|_E > \|\hat{x}_{t-1} - x_t\|_E$. One out of every 30 points are plotted to improve visibility, and results from the seen and unseen region are plotted separately.}
      \label{single_frame_reg_both}
      \vspace*{-2mm}
\end{figure}

\section{Results}
On a laptop PC with a 2.70 GHz CPU, the tracking loop ran at an average of 47.7 Hz, while the bronchoscope receives images at a rate of 25-30 Hz.

\vspace*{-1mm}
\subsection{OffsetNet Tracks Test Sequence in Seen Region} \label{seen_section}
We trained OffsetNet on $( \mathbf{I}_{x_t \in S}^{\text{cam}} , \mathbf{I}_{\hat{x}_{t-1}}^{\text{sim}} )$ image pairs from conserved regions in the left lung phantom, denoted the ``seen'' region $S$. To study the contribution of image style, we also trained a model on $( \mathbf{I}_{x_t \in S}^{\text{sim}} , \mathbf{I}_{\hat{x}_{t-1}}^{\text{sim}} )$ image pairs rendered in the same positions and orientations $x_t$. Both models were evaluated on a held-out test sequence that passed through the conserved regions of the lung phantom on which the model had been trained (seen region) and through a less conserved region that had no training images (unseen region), shown in Fig. \ref{training_dist}.

In Fig. \ref{single_frame_reg_both}, the models were analyzed in a single-frame registration task, where each $\mathbf{I}_{x_t}$ of the test sequence was tested against 25 uniformly offset $\mathbf{I}_{\hat{x}_{t-1}}$ and the results are shown separately for the seen and unseen regions of the lung. The $\mathbf{I}_{\hat{x}_{t-1}}$ were uniformly offset by 0-10 mm in $e_p$, 0-43\degree \ in $e_d$, 0-43\degree \ in $e_r$. This range covers beyond what the models were trained on to show their limits. Fig. \ref{single_frame_reg_both} visualizes this result by combining translation (mm) and rotation (\degree) offsets into a single distance measure, defined in Table \ref{notation_table}, which mimics the training loss function. When the updated location error is greater than the initial location error, OffsetNet diverges from the location estimate. Because OffsetNet's performance is correlated with the initial location error, diverging estimates increase the likelihood of failure. The results are reported in Table \ref{single_frame_table} as models $a-d$ for each dimension ($e_p$, $e_d$, $e_r$).

\begin{table}[thpb]
\caption{Results of single-frame registration task on a dataset of ($\mathbf{I}_{x_t}^{\text{cam}}$,  $\mathbf{I}_{\hat{x}_{t-1}}^{\text{sim}}$) image pairs along the test sequence. Errors reported as mean $\pm$ standard deviation.}
\centering
  \begin{tabular}{m{0.01cm} >{\centering\arraybackslash\hspace{-1pt}}m{1.5cm}  >{\centering\arraybackslash}m{0.8cm}  >{\centering\arraybackslash}m{0.9cm} >{\centering\arraybackslash} m{0.9cm}  >{\centering\arraybackslash}m{1.2cm} }
    \hline
    \multicolumn{2}{>{\centering\arraybackslash}m{1.7cm}}{\textbf{Trained On Image Styles:}}&\textbf{$\mathbf{e_p}$ [mm]}& $\mathbf{e_d}$ \newline \hspace*{-3mm}[\degree] &\textbf{$\mathbf{e_r}$}\newline \hspace*{-3mm}[\degree] &\textbf{Converging Estimates}\\
    \hline
    \rowcolor[HTML]{EFEFEF}
    \multicolumn{6}{c}{\textit{Trained in seen region, Tested in seen region}}
    \\
    $a$&$(\mathbf{I}^{\text{cam}}_{x_t \in S}, \mathbf{I}^{\text{sim}}_{\hat{x}_{t-1}})$&2.9$\pm$3.3&4.2$\pm$6.6&4.3$\pm$7.6&94\% \\ 

    $b$&$(\mathbf{I}^{\text{sim}}_{x_t \in S}, \mathbf{I}^{\text{sim}}_{\hat{x}_{t-1}})$&3.7$\pm$3.3&5.6$\pm$6.2&4.4$\pm$7.5&91\% \\ 
    \hline
    \rowcolor[HTML]{EFEFEF}
    \multicolumn{6}{c}{\textit{Trained in seen region, Tested in unseen region}}
    \\
    $c$&$(\mathbf{I}^{\text{cam}}_{x_t \in S}, \mathbf{I}^{\text{sim}}_{\hat{x}_{t-1}})$&14.5$\pm$9.3&20.5$\pm$13.0&23.4$\pm$15.4&9\% \\ 

    $d$&$(\mathbf{I}^{\text{sim}}_{x_t \in S}, \mathbf{I}^{\text{sim}}_{\hat{x}_{t-1}})$&10.2$\pm$5.9&20.6$\pm$12.0&22.4$\pm$15.0&11\% \\ 
    \hline
    \rowcolor[HTML]{EFEFEF}
    \multicolumn{6}{c}{\textit{Trained in seen and unseen region, Tested in unseen region}}
    \\
    $e$&$(\mathbf{I}^{\text{sim}}_{x_t}, \mathbf{I}^{\text{sim}}_{\hat{x}_{t-1}})$&3.8$\pm$2.3&9.2$\pm$7.6&6.2$\pm$8.9&80\%\\

    $f$&$(\mathbf{I}^{\text{rnd}}_{x_t}, \mathbf{I}^{\text{sim}}_{\hat{x}_{t-1}})$&3.2$\pm$2.4&7.5$\pm$7.7&6.0$\pm$8.6&88\%\\ 
    $g$&$(\mathbf{I}^{\text{gan}}_{x_t}, \mathbf{I}^{\text{sim}}_{\hat{x}_{t-1}})$&3.6$\pm$2.2&8.9$\pm$7.9&7.2$\pm$8.6&81\%\\ 
    \hline
  \end{tabular}
\label{single_frame_table}
\vspace*{-2mm}
\end{table}

Fig. \ref{koken_tracking} shows OffsetNet tracking the test sequence of $\mathbf{I}_{x}^{\text{cam}}$, where the output of each step updated the estimated $\hat{x}_{t}$ to be used in the following step as shown in Fig. \ref{control_loop_tracking}. The model trained on $(\mathbf{I}^{\text{cam}}_{x_t \in S}, \mathbf{I}^{\text{sim}}_{\hat{x}_{t-1}})$ image pairs tracks the bronchoscope until the bronchoscope enters the unseen region, at which point the estimate jitters around previously seen bifurcations. The maximum error in the seen region was 6.7 mm, 6.3\degree \ in $e_d$ and 5.6\degree \ in $e_r$. The model trained on $(\mathbf{I}^{\text{sim}}_{x_t \in S}, \mathbf{I}^{\text{sim}}_{\hat{x}_{t-1}})$ image pairs also failed to track in the unseen region, but surprisingly recovered from large errors in the seen region. Its maximum errors in the seen region were 29.0 mm, 16.9\degree \ in $e_d$ and 5.7\degree \ in $e_r$. 

To analyze how the tracking performances would relate to driving decisions, the visibility of airways at each location is analyzed. We define a visible airway as one whose centerline lies within 1 cm of the bronchoscope location and would lie within the location's field of view. Let $a_t$ be the set of airways visible at $x_t$, and $\hat{a_t}$ be the set of airways visible at the estimated location, $\hat{x}_t$. Sensitivity is the true positive rate, $\sum a_t \cap \hat{a}_t / \sum a_t$, while precision is the positive predictive value, $\sum a_t \cap \hat{a}_t / \sum \hat{a}_t$. Low sensitivity indicates OffsetNet missed visible airways, while low precision indicates OffsetNet misclassified airways as visible. 

Despite low tracking errors, OffsetNet trained on $(\mathbf{I}^{\text{cam}}_{x_t \in S}, \mathbf{I}^{\text{sim}}_{\hat{x}_{t-1}})$ image pairs misclassified 19\% of airways as visible, resulting mainly from the point of maximum error when it pauses at a bifurcation before continuing to track. If an autonomous agent were commanded to follow the trajectory, it is possible it would steer towards one of these mislabeled airways, which indicates this performance is insufficient for autonomous driving.


\begin{figure}[thpb]
\vspace*{-2mm}
      \centering
      \includegraphics[width=0.5\textwidth]{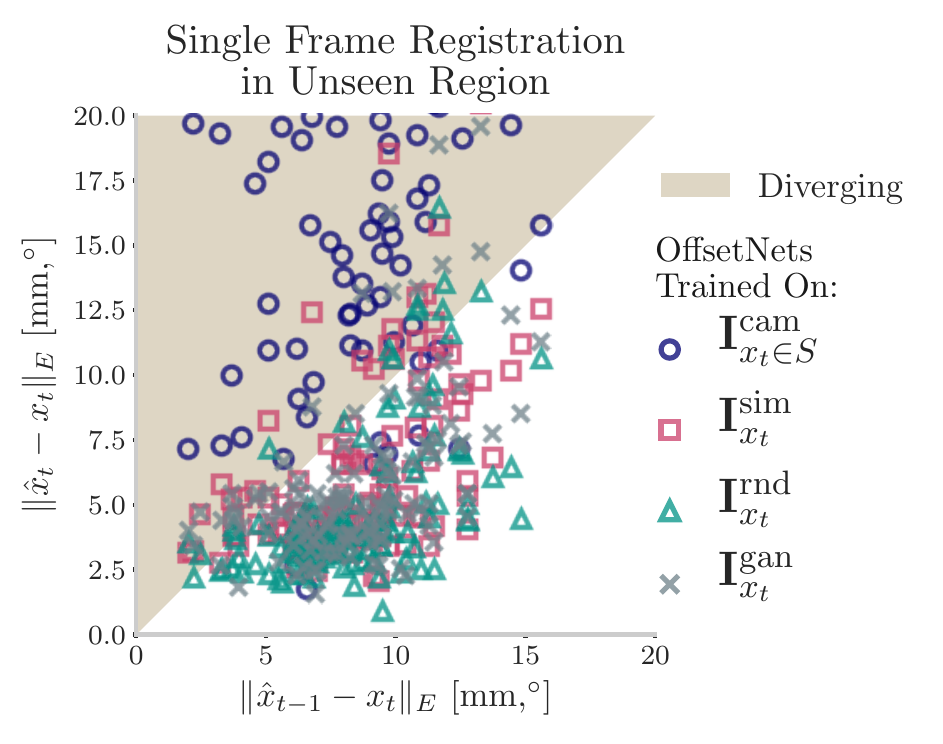}
      \caption{OffsetNets tested on 25 uniformly offset ($\mathbf{I}_{x_t \in U}^{\text{cam}}$,  $\mathbf{I}_{\hat{x}_{t-1}}^{\text{sim}}$) image pairs per $x_t$ of the test sequence in the unseen region of the lung, similar to Fig. \ref{single_frame_reg_both}. One out of every 50 points are plotted to improve visibility.}
      \label{single_frame_reg_unseen}
\end{figure}

\begin{figure*}[thpb]
      \centering
      \includegraphics[width=1.\textwidth]{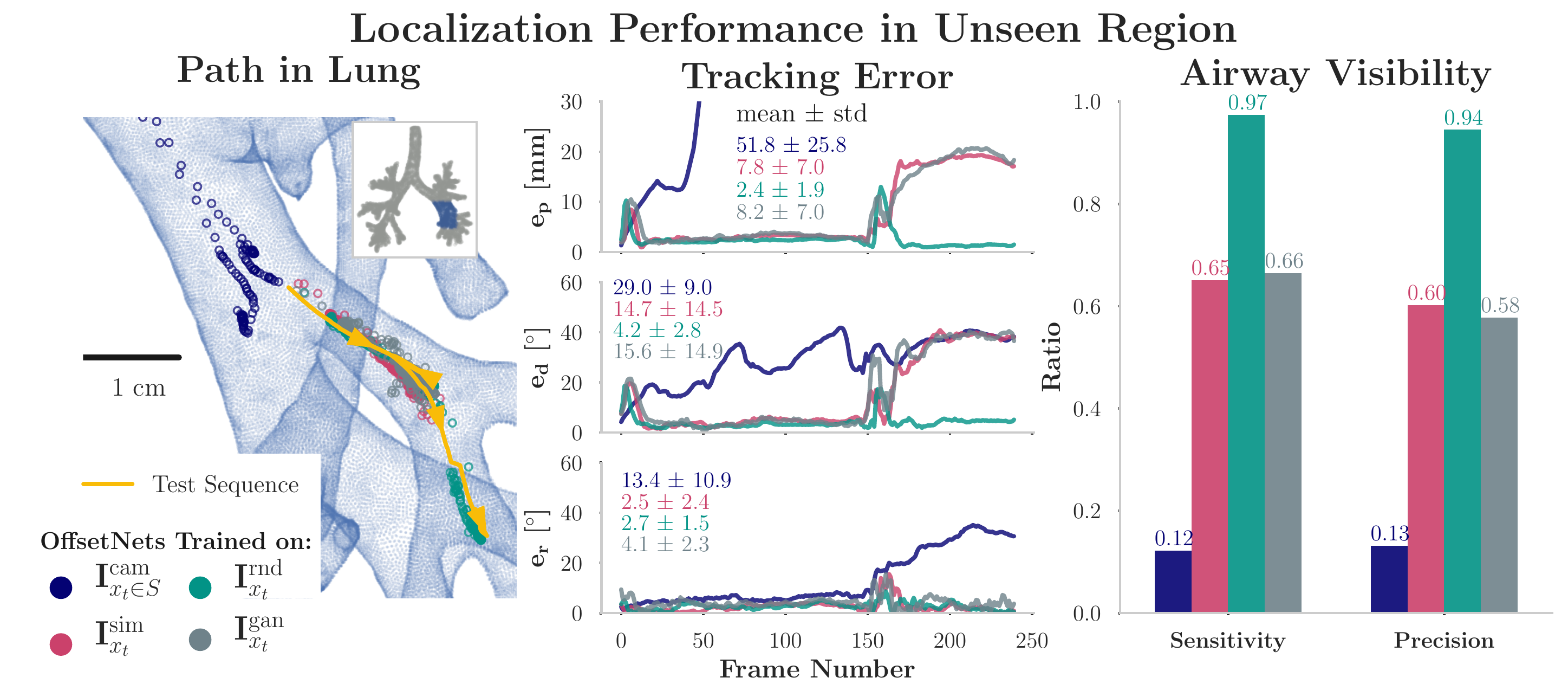}
      \caption{Similar to Fig. \ref{koken_tracking}, four OffsetNets are shown in a tracking task on a subset of the test sequence in the unseen region. Left, the path and estimated positions are shown on the lung CT. Middle, the tracking error in $e_p$, $e_d$, and $e_r$ are shown as functions of the video frame number. Right, the airway visibility analysis shows sensitivity and precision.}
      \label{tracking_unseen}
\end{figure*}

\subsection{Training with Simulated Images Enables Tracking in Unseen Region} \label{unseen_section}
In the lung periphery, the airway geometry is less conserved between patients, and the last section shows that OffsetNet's performance is extremely sensitive to unseen airway geometry \cite{Metzger2008}. We explore if training on rendered image pairs in the seen and unseen regions, $(\mathbf{I}^{\{\text{sim,rnd,gan}\}}_{x_t}$, $\mathbf{I}^{\text{sim}}_{\hat{x}_{t-1}})$,  would enable patient-specific refinement to the model. For this experiment, models are tested on $\mathbf{I}_{x \in U}^{\text{cam}}$, the subset of the images from the test sequence that lies in the unseen region $U$. Models were trained on image pairs where both $(x_t$, $\hat{x}_{t-1})$ were randomly offset around the recorded test sequence, $x_t^{\text{rec}}$, which generously targets the training around the test path. To evaluate the contribution of image style, models were trained on $(\mathbf{I}^{\{\text{sim,rnd,gan}\}}_{x_t}$, $\mathbf{I}^{\text{sim}}_{\hat{x}_{t-1}})$ image pairs, shown in Fig. \ref{lung_pics}. The three simulation-based models $e,f,g$ were compared to model $c$ trained on $(\mathbf{I}^{\text{cam}}_{x_t \in S}$, $\mathbf{I}^{\text{sim}}_{\hat{x}_{t-1}})$ image pairs. 

They were tested on the single-frame registration task, shown in Fig. \ref{single_frame_reg_unseen} and Table \ref{single_frame_table}. Models $e,f,g$ trained on simulated data outperform model $c$ on all metrics. Model $f$ trained on $(\mathbf{I}^{\text{rnd}}_{x_t}$, $\mathbf{I}^{\text{sim}}_{\hat{x}_{t-1}})$ image pairs performs the best on each metric by a small margin. The differences in performance between models $b$ and $e$ can be attributed to differences in training distribution and task difficulty.

The models were also tested on a tracking task in the unseen region, Fig. \ref{tracking_unseen}. In the tracking task, the model with no training in the region moves in the opposite direction, while all the simulation-based models track until at least the first bifurcation. Model $f$ performs the best, successfully tracking the bronchoscope into the second airway after a maximum error of 13.0 mm, 18.6\degree \ in $e_d$, an d 8.7\degree in $e_r$. Despite the high maximum error, the sensitivity and precision metrics are higher than model $a$, shown in Fig. \ref{koken_tracking}.



\section{Discussion}
The results of OffsetNet in a lung phantom demonstrate potential for being used as a real-time lung localization method. Based on only 30 minutes of training data, OffsetNet can track a held-out driving sequence in a lung phantom in real-time with accuracy comparable to or better than results reported in the literature, to the authors' knowledge. While the performance on the lung phantom may not be accurate enough to use in closed-loop control with an autonomous operator, we found that OffsetNet is capable of closing the loop in a simulated driving environment with sufficient training data; however, these results are not shown here due to space constraints.

Training on simulated data enabled tracking in a region of the lung phantom, which is an impressive result given the challenges in domain adaption for RGB cameras. This result is promising for handling airways unique to the patient in less conserved regions, given an accurate CT segmentation. While training on $(\mathbf{I}^{\text{rnd}}_{x_t}$, $\mathbf{I}^{\text{sim}}_{\hat{x}_{t-1}})$ image pairs performed better than $(\mathbf{I}^{\text{gan}}_{x_t}$, $\mathbf{I}^{\text{sim}}_{\hat{x}_{t-1}})$ image pairs here, there is room to combine both methods, and refining the GAN performance would improve OffsetNet's performance. 

OffsetNet has the potential to incorporate more information by introducing additional $\mathbf{I}_{\hat{x}_{t-1}}$ images into the tracking loop. For example, the predicted view from the centerline given the insertion information can help prevent the model from getting stuck. Multiple $\mathbf{I}_{\hat{x}_{t-1}}$ images can also provide some insight into how much confidence the model has in its predictions by measuring the variance of the resulting estimates. To maintain real-time rates, the code can be optimized by moving from Python to C++.

The algorithm has several limitations, including geometry sensitivity and tracking unpredictability. OffsetNet fails to track in airways missing from its training set. Based on the model's performance in the seen and unseen regions, it appears that OffsetNet fails to generalize to unseen airways, which emphasizes the need for creating complete training sets. OffsetNet's tracking performance is hard to predict as evidenced by the performance in Fig. \ref{tracking_unseen} when model $f$ can recover from an error, while the other models do not. More generally, stability is a risk for the tracking loop as it is only a matter of time before the location estimates $\hat{x}$ drift (Fig. \ref{control_loop_tracking}). In our tracking analysis, we strive to relate quantitative errors to the quality of the localization module, but it is imperfect. The definition of visible airways is somewhat arbitrary and may affect the reported results. In this paper, we only trained and tested OffsetNet on a limited dataset of a single lung phantom, which is considerably easier than the task of live human lungs. Finally, regardless of the algorithm, visual-based localization may struggle in the periphery of the lung when vision is lost because of airway collapse. Irrigation and air insufflation can help mitigate this issue, but there are times where the user needs to drive without vision.



In conclusion, this technique shows promise as an accurate, real-time localization method in the lung's airways. This same technique may be applied to other organ systems as well, given a suitable model of the geometry.




\section*{APPENDIX}
OffsetNet is most similar to the inverse-compositional registration technique (IC), and the table below compares the reported values in a single-frame registration task to OffsetNet \cite{Merritt2013}. Note that the distribution of images differs from the single-frame registration tasks reported above.

\begin{table}[h]
\vspace*{3mm}
\caption{Median errors for single-frame registration under random initial pose
perturbations. The errors and success rate are defined in \cite{Merritt2013}. Results are shown for OffsetNet trained on $(\mathbf{I}^{\text{cam}}_{x_t \in S}, \mathbf{I}^{\text{sim}}_{\hat{x}_{t-1}})$  image pairs [$a$] and $(\mathbf{I}^{\text{sim}}_{x_t \in S}, \mathbf{I}^{\text{sim}}_{\hat{x}_{t-1}})$ image pairs [$b$], defined in Table \ref{single_frame_table}, evaluated in the seen region.}
\centering
\vspace*{7mm}
  \caption*{(a) Tested on $(\mathbf{I}^{\text{sim}}_{x_t \in S}, \mathbf{I}^{\text{sim}}_{\hat{x}_{t-1}})$ image pairs}
  \begin{tabular}{ | >{\centering\arraybackslash}m{1.75cm} || c | c | c |  >{\centering\arraybackslash}m{1.5cm} | }
    \hline
    Method & $e_p$ & $e_d$ & $e_r$ &Success rate\\ \hline
    IC \cite{Merritt2013} & 1.6 mm & $3.1\degree$ & $1.7\degree$ & 84\% \\ \hline
    OffsetNet [$a$] & 2.4 mm & $3.4\degree$ & $2.0\degree$ & 90.2\% \\ \hline
    OffsetNet [$b$] & 1.1 mm & $1.2\degree$ & $0.9\degree$ & 99.4\% \\ \hline
  \end{tabular}
\end{table}
\begin{table}[h]
\centering
  \caption*{(b) Tested on $(\mathbf{I}^{\text{cam}}_{x_t \in S}, \mathbf{I}^{\text{sim}}_{\hat{x}_{t-1}})$ image pairs}
  \begin{tabular}{ | >{\centering\arraybackslash}m{1.75cm} || c | c | c |  >{\centering\arraybackslash}m{1.5cm} | }
    \hline
    Method & $e_p$ & $e_d$ & $e_r$ &Success rate\\ \hline
    IC \cite{Merritt2013} & 2.7 mm & $5.4\degree$ & $8.2\degree$ & 70\% \\ \hline
    OffsetNet [$a$] & 1.2 mm & $1.2\degree$ & $1.1\degree$ & 99.3\% \\ \hline
    OffsetNet [$b$] & 2.0 mm & $3.1\degree$ & $1.4\degree$ & 91.0\% \\ \hline
  \end{tabular}
\end{table}

\section*{ACKNOWLEDGMENT}
I'd like thank Auris Health Inc. for the equipment and support, NVIDIA for the Titan X GPU.
\newpage


\bibliographystyle{./IEEEtran}
\bibliography{./IEEEabrv,./library}



\end{document}